\documentclass[12pt]{article}
\usepackage[dvips]{color,graphicx}
\usepackage{amsmath,amssymb}
\usepackage{epsf}
\usepackage{natbib}

\newtheorem{theorem}{Theorem}

\newtheorem{example}[theorem]{Example}

\newtheorem{definition}[theorem]{Definition}

\title{Effects of Additional Data \\ on Bayesian Clustering}

\author{Keisuke Yamazaki\\
       k.yamazaki@aist.go.jp \\
       Artificial Intelligence Research Center,\\
       National Institute of Advanced Industrial Science and Technology\\
       2-3-26 Aomi Koto-ku, Tokyo, Japan
	}
\date{}
\begin{document}
\maketitle

\begin{abstract}
Hierarchical probabilistic models, such as mixture models, are used for cluster analysis.
These models have two types of variables: observable and latent.
In cluster analysis, the latent variable is estimated,
and it is expected that additional information will improve the accuracy of the estimation of the latent variable.
Many proposed learning methods are able to use additional data; these include semi-supervised learning and transfer learning.
However, from a statistical point of view,
a complex probabilistic model that encompasses both the initial and additional data might be less accurate
due to having a higher-dimensional parameter.
The present paper presents a theoretical analysis of the accuracy of such a model and clarifies
which factor has the greatest effect on its accuracy, the advantages of obtaining additional data,
and the disadvantages of increasing the complexity.
\end{abstract}

\section{Introduction}
\sloppy
Hierarchical probabilistic models, such as mixture models, are often used for data analysis.
These models have two types of variables: observable and latent.
Observable variables represent the data that can be observed,
while latent variables represent the hidden processes that generate the data.
In cluster analysis, for example,
the observable variable represents for the position of the observed data,
and the latent variable provides a label that indicates from which cluster
a given data point was generated.

Because there are two variables,
there are also two estimation tasks.
Prediction of the unseen data corresponds to estimating the observable variable.
Studies have performed theoretical analyses of its accuracy, and the results have been used
to determine the optimal model structure, such as by using an information criterion \citep{IEEE:Akaike:1974,watanabe_WAIC}.
On the other hand, there has not been sufficient analysis of the theoretical accuracy of estimating a latent variable.
Recently, an error function that measures the accuracy has been defined, based on the Kullback-Leibler divergence,
and an asymptotic analysis has shown that the two estimation tasks have different properties \citep{Yamazaki14a};
although when estimating the latent variable, a Bayesian clustering method is more accurate
than the maximum-likelihood clustering method,
these two methods have the same asymptotic error in a prediction task.

In practical applications of cluster analysis, increasing the accuracy is one of the central issues.
For this reason, clustering based on additional information is promising,
and many methods have been developed that include additional prior knowledge 
in the model structure; examples include semi-supervised learning \citep{Zhu:Survey,Chapelle+2006}, transductive learning \citep{Chapelle99}, 
transfer learning, and domain adaptation \citep{Raina05}.
In multitask learning \citep{Caruana97,Marx05,Raina05},
there are various classification tasks, and the goal is to solve them simultaneously.
Although multitask learning is symmetric in terms of the importance of each task,
in the present paper, we focus on estimation tasks that are asymmetric:
there is an initial data set that presents the primary task 
and an additional data set that is not estimated but supplies supplementary information \citep{Kaski07}.
This type of task includes semi-supervised learning, transductive learning, and transfer learning.

From a statistical point of view, the degree to which the additional data improves the estimation is not trivial.
The structure of the model will be more complex and thus able to accept additional data.
There is a trade-off between the complexity of a model and the amount of data that is used;
a more complex model will be less accurate due to the increased dimensionality of the parameter,
but more data will improve the accuracy.
In a prediction task, where the estimation target is the observable variable,
it has been proven mathematically that the advantages of increasing the amount of data outweighs the disadvantages of increasing the complexity of the model \citep{Yamazaki&Kaski09}.
However, it is a still open question whether the use of additional data increases clustering accuracy in methods other than semi-supervised learning, where this has already been proven \citep{Yamazaki15c,Yamazaki15a}.

In the present paper, we extend the results of \citet{Yamazaki15c,Yamazaki15a}, and investigate the effect of additional data on the accuracy of Bayesian clustering.
We consider a mixture model that uses both initial and additional data.
When the additional data are ignored, only the initial data are used to determine the structure of the model, such as the dimensionality of the data and the number of clusters,  
and thus the dimensionality of the parameter decreases; that is, the model becomes less complex.
By comparing the accuracy with and without the use of additional data,
we clarify the effect of the additional data in the asymptotic case, that is,
when the total amount initial and additional data is sufficiently large.
Moreover, the extension of the present paper allows us to elucidate the effect of more complicated overlap
between the initial and the additional data sets.
For example, we will deal with the unlabeled additional data while the former study \citep{Yamazaki15c} restricts the analysis to the labeled ones.

The remainder of this paper is organized as follows.
Section \ref{sec:Bayes} summarizes Bayesian clustering and considers its asymptotic accuracy when there are  no additional data.
Section \ref{sec:Def_Data} presents a formal definition of a mixture model that incorporates additional data
and derives the asymptotic accuracy of the model.
Section \ref{sec:Effective_Data} determines under what conditions the use of additional data will improve the accuracy.
Finally, we present a discussion and our conclusions in Sections \ref{sec:Dis} and \ref{sec:Con}, respectively.
\section{Bayesian Clustering}
\label{sec:Bayes}
In this section, we present a definition of Bayesian clustering
and present an evaluation function for the clustering results.

We consider a mixture model defined by
\begin{align*}
p(x|w) =& \sum_{k=1}^K a_k f(x|b_k),
\end{align*}
where $x \in R^d$ expresses the position of a data point,
$w$ is the parameter, 
and $f$ is the density function associated with a mixture component.
The mixing ratio $a_k$ has constraints given by
$a_k>0$ for all $k$, and $a_K=1-\sum_{k=1}^{K-1} a_k$.
Let the dimension of $b_k$ be $\dim b_k =d_c$,
that is, $b_k=(b_{k1},\dots,b_{kd_c})^\top$.
Then, the parameter $w$ can be expressed as
\begin{align*}
w =& (a_1,\dots,a_{K-1},b_{11},\dots,b_{1d_c},\dots,b_{Kd_c})^\top.
\end{align*}

We define the data source, which generates i.i.d.\ data, as follows:
\begin{align*}
q(x,y) =& q(y)q(x|y) = a^*_y f(x|b^*_y),
\end{align*}
where $y$ indicates a cluster label,
and $a^*_y$ and $b^*_y$ are constants.
So that the clusters can be identified, we require $b^*_i\ne b^*_j$ for $i\ne j$.
The data source is $q(x,y)=p(x,y|w^*)$,
where $p(x,y|w)=a_y f(x|b_y)$ and 
\begin{align*}
w^* = (a^*_1,\dots, a^*_{K-1},b^*_{11},\dots,b^*_{Kd_c})^\top.
\end{align*}
Note that identification of the labels $i$ and $j$ are impossible in the unsupervised learning
when the components have the same parameter $b^*_i=b^*_j$.
We refer to $w^*$ as the true parameter.
Let $(X^n,Y^n)=\{(x_1,y_1),\dots,(x_n,y_n)\}$ be generated by the data source.
We use the notation $X^n=\{x_1,\dots,x_n\}$ and $Y^n=\{y_1,\dots,y_n\}$
for the sets of data positions and labels, respectively.

Cluster analysis is formulated as estimating $Y^n$ when $X^n$ is given.
When considered as a density estimation, the task is to estimate $p(Y^n|X^n)$.
If $Y^n$ is observable, the task must be to estimate unseen $x$.
This is the prediction of the supervised learning and its theoretical analysis has been thoroughly studied
\citep{IEEE:Akaike:1974,Watanabe01a}.
Since the label $Y^n$ is not given, the latent variable explicitly appears in the clustering algorithms
such as the expectation-maximization algorithm \citep{Dempster77} and the variational Bayes algorithm \citep{Attias99}.

Bayesian clustering is then defined as
\begin{align*}
p(Y^n|X^n) =& \int \prod_{i=1}^n p(y_i|x_i,w) p(w|X^n)dw,
\end{align*}
where the conditional probability $p(y|x,w)$ is defined as
\begin{align*}
p(y|x) =& \frac{p(x,y|w)}{p(x|w)} = \frac{a_y f(x|b_y)}{p(x|w)},
\end{align*}
and $p(w|X^n)$ is the posterior distribution. 
When a prior distribution is given by $\varphi(w)$,
the posterior distribution is defined as
\begin{align*}
p(w|X^n) =& \frac{1}{Z(X^n)}\prod_{i=1}^n p(x_i|w)\varphi(w),
\end{align*}
where $Z(X^n)$ is the normalizing factor
\begin{align*}
Z(X^n) =& \int \prod_{i=1}^n p(x_i|w)\varphi(w) dw.
\end{align*}
If we replace $p(w|X^n)$ in $p(Y^n|X^n)$ with this definition,
we find an equivalent expression for the estimated density:
\begin{align*}
p(Y^n|X^n) =& \frac{\int \prod_{i=1}^n p(x_i,y_i|w)\varphi(w)dw}
{\int \prod_{i=1}^n p(x_i|w) \varphi(w)dw}.
\end{align*}

Since the clustering task is formulated as a density estimation,
the difference between the true density of $Y^n$ and the estimated density, $p(Y^n|X^n)$, can be used to
evaluate the accuracy of the clustering.
The true density is defined as
\begin{align*}
q(Y^n|X^n) =& \frac{q(X^n,Y^n)}{q(X^n)} 
= \prod_{i=1}^n \frac{q(x_i,y_i)}{\sum_{y_i=1}^K q(x_i,y_i)}.
\end{align*}
In the present paper, we will use the Kullback-Leibler divergence
to measure the difference between the densities:
\begin{align*}
D(n) =& \frac{1}{n}E_{X^n}\bigg[\sum_{Y^n} q(Y^n|X^n)\ln \frac{q(Y^n|X^n)}{p(Y^n|X^n)}\bigg],
\end{align*}
where $E_{X^n}[\cdot]$ is the expectation over all $X^n$.
We evaluate the density estimation of $Y^n$ since $Y^n$ is the probabilistic variable due to the generating process of the data source.
This is the reason why we consider the divergence instead of the deterministic loss function
such as the 0-1 loss.

We wish to find the asymptotic form of the error function $D(n)$, that is, the case in which the number of data points $n$ is sufficiently large.
Assume that the Fisher information matrices
\begin{align*}
\{ I_{XY}(w^*) \}_{ij} =& E\biggl[ \frac{\partial \ln p(x,y|w^*)}{\partial w_i}
\frac{\partial \ln p(x,y|w^*)}{\partial w_j}\biggr],\\
\{ I_{X}(w^*) \}_{ij} =& E\biggl[ \frac{\partial \ln p(x|w^*)}{\partial w_i}
\frac{\partial \ln p(x|w^*)}{\partial w_j}\biggr]\\
\end{align*}
exist and are positive definite,
where the expectation is
\begin{align*}
E[f(x,y)]=\int \sum_{y=1}^{K} f(x,y)p(x,y|w^*)dx.
\end{align*}
This assumption corresponds to the statistical regularity,
which requires that there is no redundant component of the model compared with the data source.
In the case, where the regularity is not satisfied,
the algebraic geometrical analysis is available \citep{Watanabe01a,Yamazaki2016}.
The present paper focuses on the regular case.
We then have the following theorem \citep{Yamazaki14a}.
\begin{theorem}
\label{th:asymD}
The error function $D(n)$ has the asymptotic form
\begin{align*}
D(n) =& \frac{1}{2}\ln\det \big[ I_{XY}(w^*)I_X(w^*)^{-1}\big] \frac{1}{n} 
+ o\bigg(\frac{1}{n}\bigg).
\end{align*}
\end{theorem}
Since the data source is described by the model,
the posterior distribution converges to the true parameter.
Then, the error goes to zero in the asymptotic limit.
This theorem shows the convergence speed;
the leading term has the order $1/n$, and its coefficient is determined by
the Fisher information matrices.
\section{Formal Definition of an Additional Data Set}
\label{sec:Def_Data}
In this section, we formally define an additional data set
and perform a clustering task for a given data set.
We then derive the asymptotic form of the error function.

\subsection{Formulation of Data and Clustering}
Let the initial data set be denoted $D_i$, that is, $D_i=X^n$.
Let an additional data set be denoted $D_a$.
We assume that the additional data comprise $n$ ordered elements:
\begin{align*}
D_a = \{ z_{n+1},\dots,z_{n+\alpha n} \},
\end{align*}
where $\alpha$ is positive and real, and $\alpha n$ is an integer.
The element $z_i$ is $x_i$ for an unlabeled case, and it
is $(x_i,y_i)$ for a labeled case.
Let $p_a(z|v)$ be the density function of $z$,
where $v$ is the parameter.
Assume that the data source of the additional data is defined by
\begin{align*}
q_a(z) = p_a(z|v^*),
\end{align*}
where a constant $v^*$ is the true parameter. Also, assume
that the following Fisher information matrix exists and is positive definite:
\begin{align*}
\{I_Z(v^*)\}_{ij} =& E_z\bigg[\frac{\partial \ln p_a(z|v^*)}{\partial v_i}
\frac{\partial \ln p_a(z|v^*)}{\partial v_j}\bigg],
\end{align*}
where the expectation is based on $p_a(z|v^*)$;
for the unlabeled case
\begin{align*}
E_z\big[ f(z) \big] =& \int f(x)p_a(x|v^*)dx,
\end{align*}
and for the labeled case,
\begin{align*}
E_z\big[ f(z) \big] =& \int \sum_y f(x,y)p_a(x,y|v^*)dx.
\end{align*}

Let us consider a parameter vector $u$ defined by
\begin{align*}
u =& (u_1,\dots,u_{d_1},u_{d_1+1},\dots,u_{d_1+d_2},u_{d_1+d_2+1},\dots,u_{d_1+d_2+d_3})^\top.
\end{align*}
This parameter is divided into three parts:
$(u_1,\dots,u_{d_1})$ contains the elements 
included in $w$ but not in $v$,
$(u_{d_1+1},\dots,u_{d_1+d_2})$ contains the elements 
included in both $w$ and $v$,
and $(u_{d_1+d_2+1},\dots,u_{d_1+d_2+d_3})$ contains the elements 
included in $v$ but not in $w$.
This means that there are permutations $\psi_i$ and $\psi_a$ defined by
\begin{align*}
(u_1,\dots,u_{d_1},u_{d_1+1},\dots,u_{d_1+d_2})^\top =& \psi_i(w),\\
(u_{d_1+1},\dots,u_{d_1+d_2},u_{d_1+d_2+1},\dots,u_{d_1+d_2+d_3})^\top =& \psi_a(v),
\end{align*}
which are sorting functions for $w$ and $v$, respectively.
Considering these permutations,
we use the notation $p_i(x|u)=p(x|w)$ and $p_i(x,y|u)=p(x,y|w)$ 
for the initial data set,
and $p_a(x|u)=p_a(x|v)$ and $p_a(x,y|u)=p_a(x,y|v)$ for the additional data set.

Bayesian clustering with an additional data set is defined as
\begin{align*}
p(Y^n|X^n,D_a) =& \frac{\int \prod_{j=1}^n p_i(x_j,y_j|u)\prod_{i=n+1}^{n+\alpha n}p_a(z_i|u)\varphi(u) du}
{\int \prod_{i=1}^n p_j(x_j|u)\prod_{i=n+1}^{n+\alpha n}p_a(z_i|u)\varphi(u) du},
\end{align*}
where the prior distribution is $\varphi(u)$.
Note that the estimation target is $Y^n$,
and labels are not estimated for the additional data,
even if they are unlabeled.
\subsection{Four Cases of Additional Data Sets}
According to the division of the parameter dimension $d_1$, $d_2$ and $d_3$,
we will consider the following four cases:
\begin{enumerate}
\item $d_1=d_3=0$, 
\item $d_1>0,d_3=0$, 
\item $d_1=0,d_3>0$,
\item $d_1>0,d_3>0$,
\end{enumerate}
where $d_2>0$ is assumed in the all cases.
Note that the additional data do not affect the clustering result when $d_2=0$
since there is no overlap between the models for the initial and the additional data.

\begin{figure}[tp]
\begin{tabular}{c|c}
\begin{minipage}{0.5\hsize}
\centering
\includegraphics[angle=-90,width=\columnwidth]{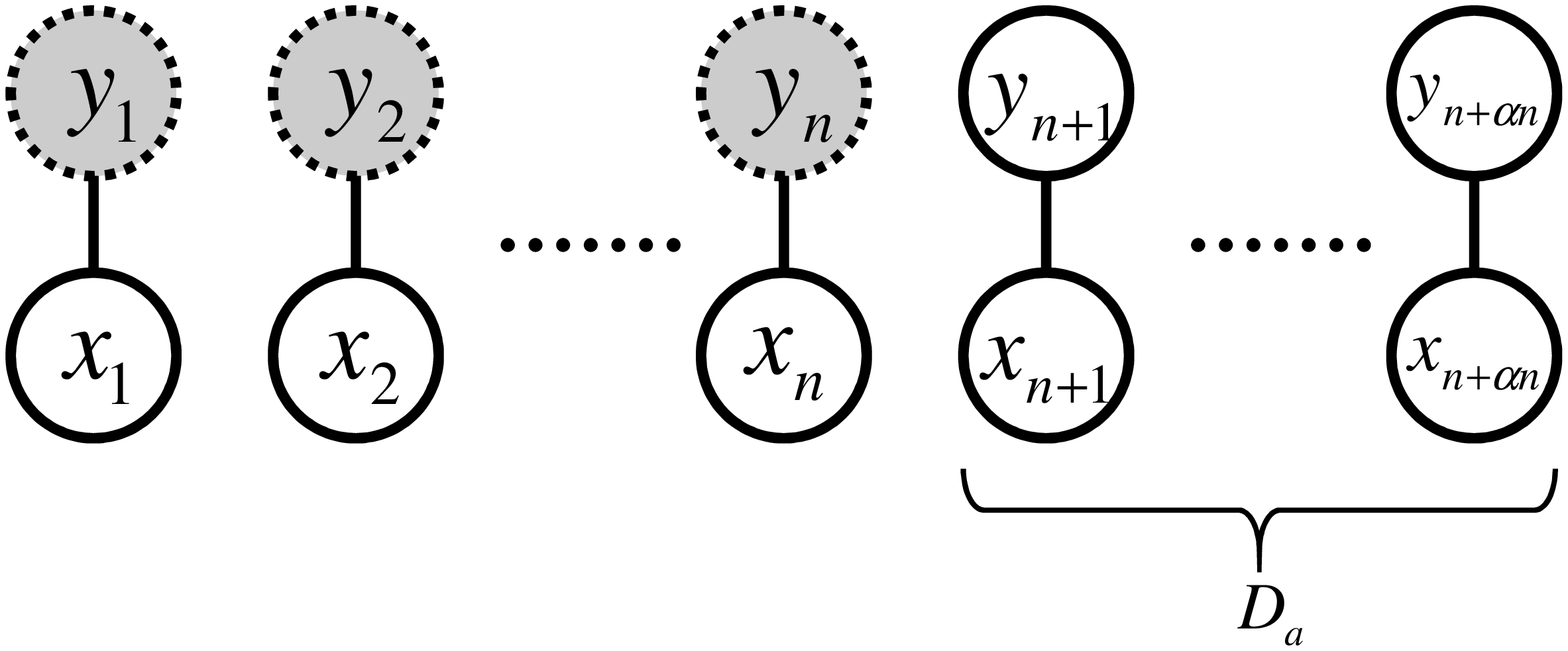}
\text{Example 2}
\label{fig:ex2}
\end{minipage}
&
\begin{minipage}{0.5\hsize}
\centering
\includegraphics[angle=-90,width=\columnwidth]{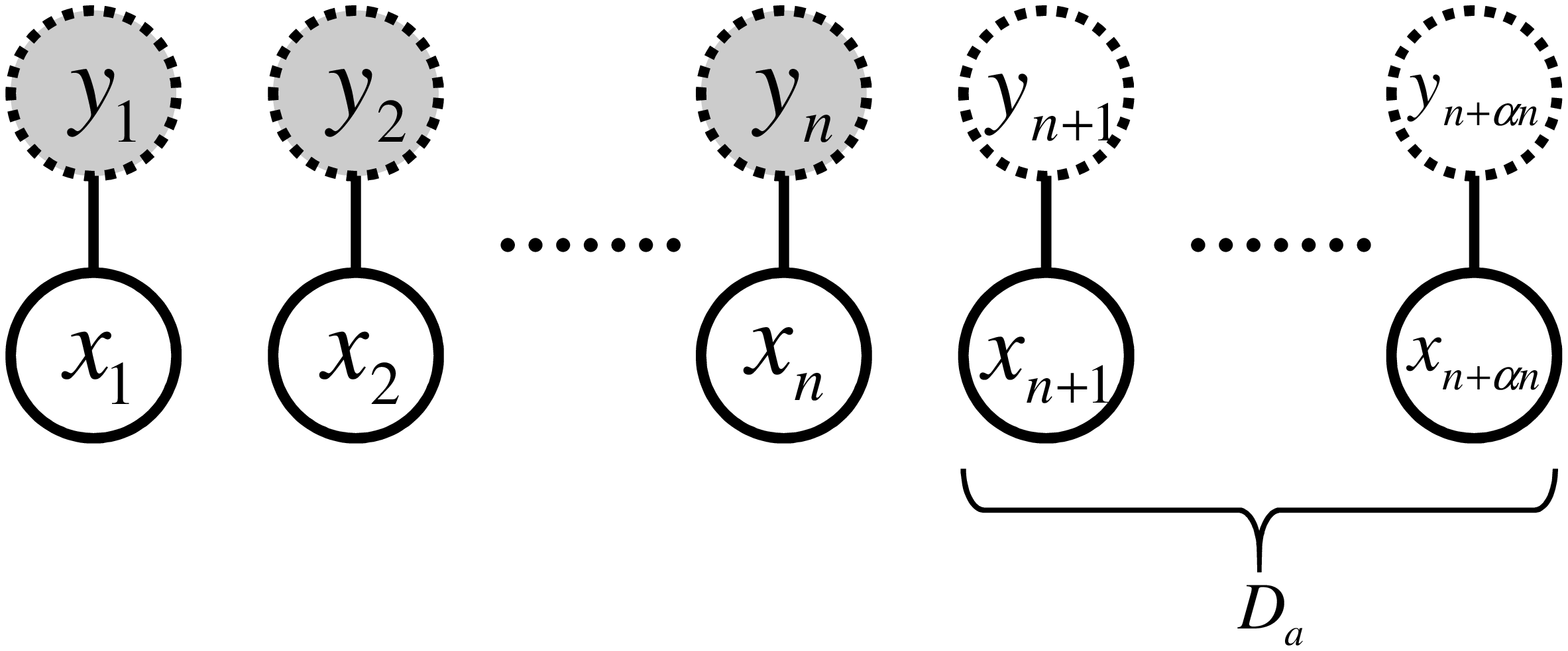}
\text{Example 3}
\label{fig:ex3}
\end{minipage}
\\
\hline
\begin{minipage}{0.5\hsize}
\centering
\includegraphics[angle=-90,width=\columnwidth]{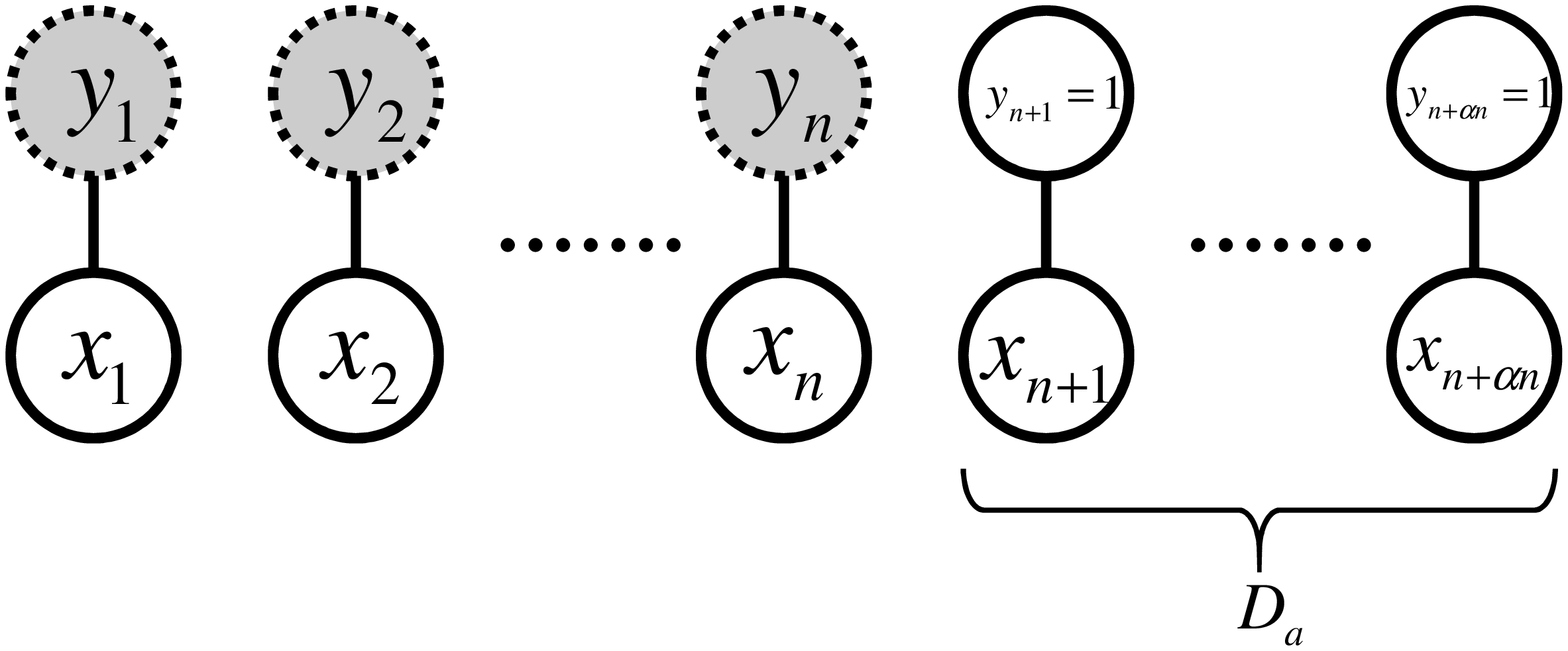}
\text{Example 4}
\label{fig:ex4}
\end{minipage}
&
\begin{minipage}{0.5\hsize}
\centering
\includegraphics[angle=-90,width=\columnwidth]{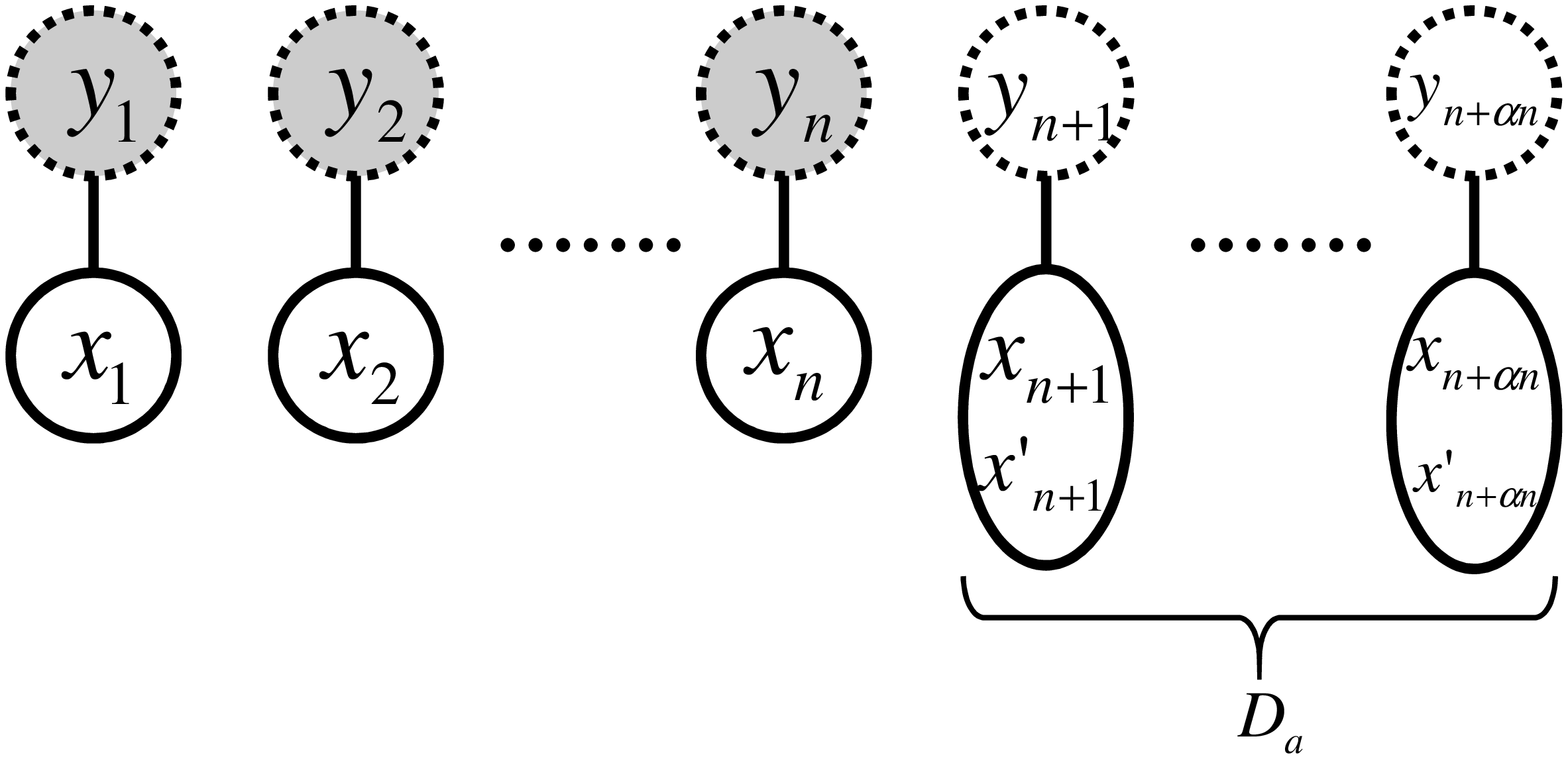}
\text{Example 5}
\label{fig:ex5}
\end{minipage}
\\
\hline
\begin{minipage}{0.5\hsize}
\centering
\includegraphics[angle=-90,width=\columnwidth]{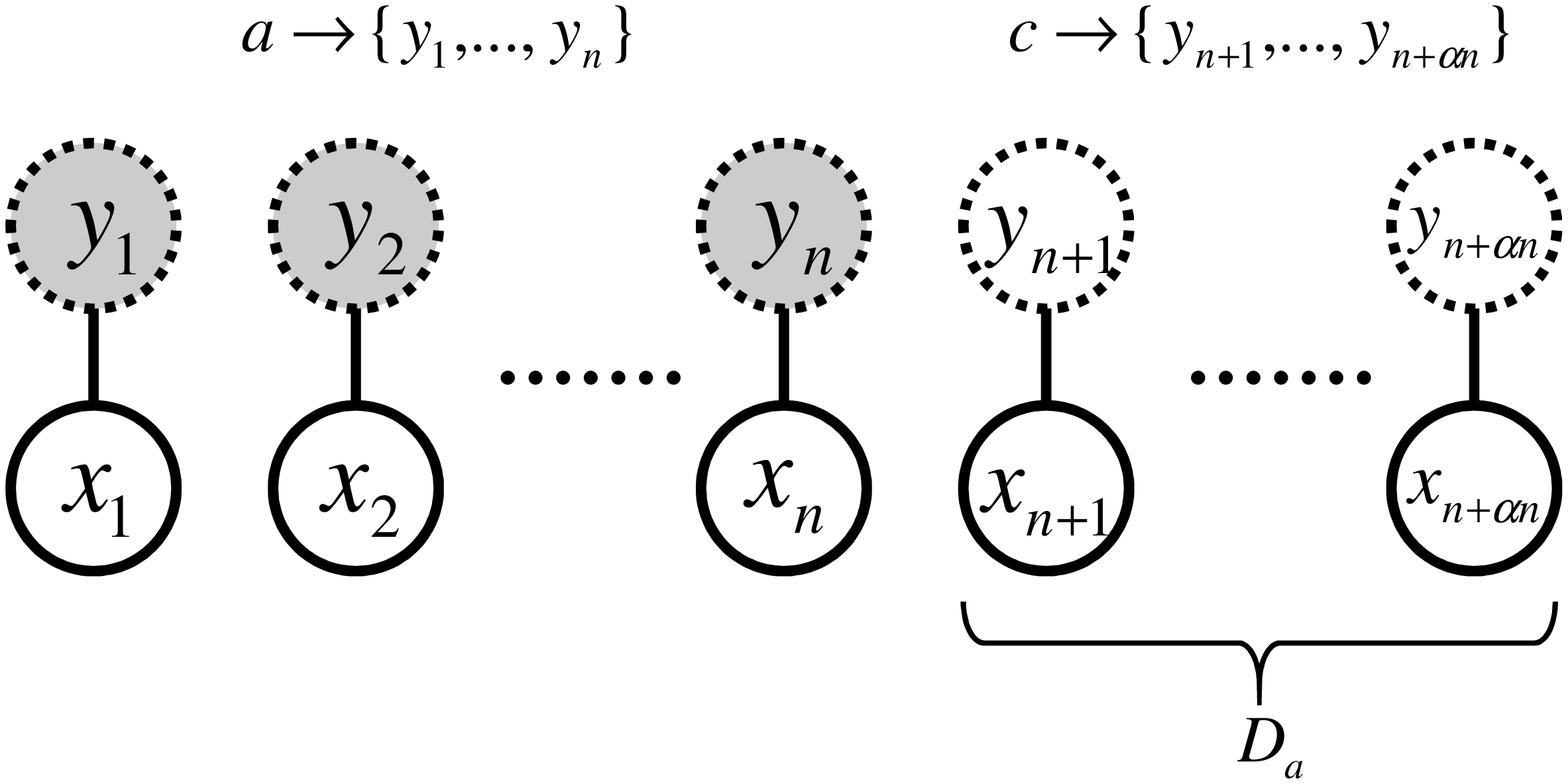}
\text{Example 6}
\label{fig:ex6}
\end{minipage}
&
\end{tabular}
\caption{Estimations of latent variables when using additional data sets.
The initial data are $\{x_1,\dots,x_n\}$. Solid and dotted nodes indicate the observable and unobservable variables, respectively.
Gray nodes are the estimation targets. $D_a$ indicates the additional data set.}
\label{fig:examples}
\end{figure}
The following two examples are the first case, where $d_1=d_3=0$;
\begin{example}[Semi-supervised learning]
\label{ex:SSL}
Semi-supervised classification \citep[Type II$'$][]{Yamazaki12c}, \citep{Zhu:Survey}
is described by
\begin{align*}
p_i(x|u) =& p(x|w) = \sum_{k=1}^K a_k f(x|b_k),\\
p_a(x,y|u) =& p(x,y|w) = a_y f(x|b_y),
\end{align*}
where the parameter is given by
\begin{align*}
u =& (a_1,\dots,a_{K-1},b_{11},\dots,b_{Kd_c})^\top.
\end{align*}
In this case, $u=w=v$ and $d_1=d_3=0$.
The unlabeled data $X^n$ and the labeled data 
$D_a=\{(x_{n+1},y_{n+1}),\dots,(x_{n+\alpha n},y_{n+\alpha n})\}$
are generated by $p_i(x|u^*)$ and $p_a(x,y|u^*)$, respectively.
The clustering task is to estimate the density of $Y^n$:
\begin{align*}
p(Y^n|X^n,D_a) =& \frac{\int \prod_{i=1}^n a_{y_i}f(x_i|b_{y_i}) 
\prod_{i=n+1}^{\alpha n} a_{y_i}f(x_i|b_{y_i})\varphi(u)du}
{\int \prod_{i=1}^n \sum_{y=1}^K a_yf(x_i|b_y) 
\prod_{i=n+1}^{n+\alpha n} a_{y_i}f(x_i|b_{y_i})\varphi(u)du}.
\end{align*}
The schematic relation between the initial and the additional data sets is shown in the top-left panel of Figure \ref{fig:examples}.
\end{example}
\begin{example}
\label{ex:TypeIId}
Clustering of a partial data set \citep{Yamazaki14a} is described by
\begin{align*}
p_i(x|u) =& p_a(x|u) = \sum_{k=1}^K a_k f(x|b_k),
\end{align*}
where $u=w=v$ and $d_1=d_3=0$.
Both the initial data $X^n$ and the additional data $D_a$ are unlabeled,
which corresponds to the estimation of $n$ labels based on ($n+\alpha n$) data points:
\begin{align*}
p(Y^n|X^n,D_a) =& \frac{\int \prod_{i=1}^n a_{y_i}f(x_i|b_{y_i}) 
\prod_{i=n+1}^{n+\alpha n} \sum_{y=1}^K a_yf(x_i|b_y)\varphi(u)du}
{\int \prod_{i=1}^n \sum_{y=1}^K a_yf(x_i|b_y) 
\prod_{i=n+1}^{n+\alpha n} \sum_{y=1}^K a_yf(x_i|b_y)\varphi(u)du}.
\end{align*}
The relation between the initial and the additional data sets is shown in the top-right panel of Figure \ref{fig:examples}.
\end{example}
The next case is an example of the second case, where $d_1>0$ and $d_3=0$;
\begin{example}
\label{ex:one_class}
Suppose some cluster provides labeled data in the additional data set.
For example, suppose the labeled data of the first cluster, which is the target, are given in $D_a$.
In other words, the positive labeled data are additionally given \citep{Plessis+2015}.
Then, the density functions are defined as
\begin{align*}
p_i(x|u) =& \sum_{k=1}^K a_k f(x|b_k),\\
p_a(x|u) =& f(x|b_1),
\end{align*}
where $D_a=\{(x_{n+1},1),\dots,(x_{n+\alpha n},1)\}$.
The parameter vector is expressed by
\begin{align*}
u =& (a_1,\dots,a_{K-1},b_{21},\dots,b_{Kd_c},b_{11},\dots,b_{1d_c})^\top,
\end{align*}
where the common part is $(b_{11},\dots,b_{1d_c})$, and $d_3=0$.
The estimated density is given by
\begin{align*}
p(Y^n|X^n,D_a) =& \frac{\int \prod_{i=1}^n a_{y_i}f(x_i|b_{y_i}) 
\prod_{i=n+1}^{\alpha n} f(x_i|b_1)\varphi(u)du}
{\int \prod_{i=1}^n \sum_{y=1}^K a_yf(x_i|b_y) 
\prod_{i=n+1}^{n+\alpha n} f(x_i|b_1)\varphi(u)du}.
\end{align*}
The relation between the initial and the additional data sets is shown in the middle-left panel of Figure \ref{fig:examples}.
\end{example}
The case, where $d_1=0$ and $d_3>0$, has the following example;
\begin{example}
\label{ex:add_feature}
When a new feature $x'$ is added to $D_a$,
the density functions are defined as
\begin{align*}
p_i(x|u) =& \sum_{k=1}^K a_k f(x|b_k),\\
p_a(x,x'|u) = & \sum_{k=1}^K a_kf(x|b_k)g(x'|c_k),
\end{align*}
where $z=(x,x')$, and $x'$ is generated by $g(\cdot|c_y)$.
For simplicity, let $x'$ and $c_y$ be scalar,
and let $x'$ be conditionally independent of $x$.
Note that the asymptotic results of the present paper
hold when this assumption is not satisfied.
The parameter vector is expressed as
\begin{align*}
u =& (a_1,\dots,a_{K-1},b_{11},\dots,b_{Kd_c},c_1,\dots,c_K)^\top,
\end{align*}
where $(a_k,\dots,b_{Kd_c})$ is the common part,
$d_1=0$, and $d_3=K$.
The estimated density is given by
\begin{align*}
p(Y^n|X^n\hskip-1mm,\hskip-1mmD_a)\hskip-1mm =& \frac{\int \prod_{i=1}^n a_{y_i}f(x_i|b_{y_i}) 
\prod_{i=n+1}^{\alpha n} \sum_{y=1}^Ka_yf(x_i|b_y)g(x'_i|c_y)\varphi(u)du}
{\int \prod_{i=1}^n \sum_{y=1}^K a_yf(x_i|b_y) 
\prod_{i=n+1}^{n+\alpha n} \sum_{y=1}^Ka_yf(x_i|b_y)g(x'_i|c_y)\varphi(u)du}.
\end{align*}
The relation between the initial and the additional data sets is shown in the middle-right panel of Figure \ref{fig:examples}.
\end{example}
Lastly, the case, where $d_1,d_3>0$, has the following example;
\begin{example}
\label{ex:CPchange}
When the class-prior changes \citep{Plessis+2014,Yamazaki15c},
the density functions are described by
\begin{align*}
p_i(x|u) =& \sum_{k=1}^K a_k f(x|b_k),\\
p_a(x|u) =& \sum_{K=1}^K c_k f(x|b_k),
\end{align*}
where the mixing ratio of the additional data $c_k$ for $1\le k\le K$
is different from that of the initial data $a_k$.
In the analysis of the previous study, the additional data are restricted to the labeled ones \citep{Yamazaki15c},
which is expressed as
\begin{align*}
p_a(x,y|u) =& c_k f(x|b_k).
\end{align*}
In the present paper, we extend the situation to the unlabeled case.
The parameter vector is given by
\begin{align*}
u =& (a_1,\dots,a_{K-1},b_{11},\dots,b_{Kd_c},c_1,\dots,c_{K-1})^\top,
\end{align*}
where $(b_{11},\dots,b_{Kd_c})$ is the common part, and $d_1=d_3=K-1$.
The estimated density is given by
\begin{align*}
p(Y^n|X^n,D_a) =& \frac{\int \prod_{i=1}^n a_{y_i}f(x_i|b_{y_i}) 
\prod_{i=n+1}^{\alpha n} \sum_{y=1}^K c_yf(x_i|b_y)\varphi(u)du}
{\int \prod_{i=1}^n \sum_{y=1}^K a_yf(x_i|b_y) 
\prod_{i=n+1}^{n+\alpha n} \sum_{y=1}^K c_yf(x_i|b_y)\varphi(u)du}.
\end{align*}
The relation between the initial and the additional data sets is shown in the bottom panel of Figure \ref{fig:examples}.
The notation $a \rightarrow \{y_1,\dots,y_n\}$ and $c \rightarrow \{y_{n+1},\dots,y_{n+\alpha n}\}$
show that the latent variables $y_i$ are based on a model with a mixing ratio of $a$ to $c$.
\end{example}
Example \ref{ex:one_class} is a special case of Examples \ref{ex:SSL} and \ref{ex:CPchange}.
\subsection{Error Function with Additional Data}
In the previous studies, we formulated the error function and derived its asymptotic form in each case \cite{Yamazaki14a,Yamazaki15a,Yamazaki15c}.
Here, we show the unified formulation and derivation of the error function.
The error function is given by
\begin{align}
D_a(n) = E_{XD}\bigg[\sum_{Y^n}q(Y^n|X^n)\ln \frac{q(Y^n|X^n)}{p(Y^n|X^n,D_a)}\bigg], \label{eq:def_Da}
\end{align}
where $E_{XD}[\cdot]$ is the expectation over all $X^n$ and $D_a$.
Define three Fisher information matrices as
\begin{align*}
\{I_{XY}(u)\}_{jk} =& E\bigg[ \frac{\partial \ln p_i(x,y|u)}{\partial u_j}
\frac{\partial \ln p_i(x,y|u)}{\partial u_k} \bigg],\\
\{I_X(u)\}_{jk} =& E\bigg[ \frac{\partial \ln p_i(x|u)}{\partial u_j}
\frac{\partial \ln p_i(x|u)}{\partial u_k} \bigg],\\
\{I_Z(u)\}_{jk} =& E_z \bigg[ \frac{\partial \ln p_a(z|u)}{\partial u_j}
\frac{\partial \ln p_a(z|u)}{\partial u_k} \bigg].
\end{align*}
An asymptotic property of the error is determined by these matrices.
\begin{theorem}
\label{th:asymD_a}
The error function $D_a(n)$ has the asymptotic form
\begin{align*}
D_a(n) =& \frac{1}{2}\ln \det J_{XY}(u^*)J_X(u^*)^{-1}\frac{1}{n}+o\bigg(\frac{1}{n}\bigg),
\end{align*}
where 
\begin{align*}
J_{XY}(u) =& I_{XY}(u) + \alpha I_Z(u),\\
J_X(u) =& I_X(u) + \alpha I_Z(u).
\end{align*}
\end{theorem}
By generalizing the derivation of Theorem 2 in \citet{Yamazaki15a},
the proof is shown as follows;
\newline
{\bf Proof of Theorem \ref{th:asymD_a}:}
\newline
Based on the definition, the error function can be divided into two parts:
\begin{align*}
nD_a(n) =& F_{XY}(n) - F_X(n),
\end{align*}
where the free energy functions are given by
\begin{align*}
F_{XY}(n) =& -nS_{XY} - E_{XD}\bigg[\ln \int \prod_{j=1}^n p_i(x_j,y_j|u)
\prod_{i=n+1}^{n+\alpha n} p_a(z_i|u)\varphi(u)du\bigg],\\
F_X(n) =& -nS_X - E_{XD}\bigg[\ln \int \prod_{j=1}^n p_i(x_j|u)
\prod_{i=n+1}^{n+\alpha n} p_a(z_i|u)\varphi(u)du\bigg].
\end{align*}
The entropy functions are defined as
\begin{align*}
S_{XY} =& E\big[ - \ln p(x,y|u^*)\big],\\
S_X =& E\big[ -\ln p(x|u^*)\big].
\end{align*}
Based on the saddle point approximation and the assumptions
on the Fisher information matrices $I_{XY}(u^*)$, $I_X(u^*)$, and $I_Z(u^*)$, 
it has been shown that the free energy functions have 
the following asymptotic forms \citep{Clarke90,Yamazaki14a,Yamazaki15c}:
\begin{align*}
F_{XY}(n) =& \frac{\dim u}{2}\ln \frac{n}{2\pi e} 
+ \ln \frac{\sqrt{\det J_{XY}(u^*)}}{\varphi(u^*)} + o(1),\\
F_X(n) =& \frac{\dim u}{2}\ln \frac{n}{2\pi e} 
+ \ln \frac{\sqrt{\det J_X(u^*)}}{\varphi(u^*)} + o(1).
\end{align*}
Rewriting the energy functions in their asymptotic form,
we obtain 
\begin{align*}
nD_a(n) =& \frac{1}{2}\ln \det J_{XY}(u^*) -\frac{1}{2}\ln \det J_X(u^*) + o(1),
\end{align*}
which completes the proof.
{\bf (End of Proof)}
\section{Effective Additional Data Sets}
\label{sec:Effective_Data}
In this section, we determine when the use of an additional data set
makes the estimation more accurate.
\subsection{Formal Definition of Effective Data Set and Sufficient Condition for Effectiveness}
Using the asymptotic form of the error functions, we formulate as follows
an additional data set that improves the accuracy.
\begin{definition}[Effective data set]
If the difference between the error with and without a particular additional data set $D_a$
satisfies the following condition, then the data set is {\emph effective}:
there exists a positive constant $C$ such that
\begin{align*}
D(n) - D_a(n) =& \frac{C}{n} + o\bigg(\frac{1}{n}\bigg).
\end{align*}
\end{definition}
According to this definition, $D_a$ is effective
if the leading term of the asymptotic form of $D_a(n)$ is smaller than that of $D(n)$.

Let us rewrite $I_{XY}(u^*)$ and $I_X(u^*)$ as block matrices:
\begin{align*}
I_{XY}(u^*) =&
\begin{pmatrix}
K_{11} & K_{12} \\
K_{21} & K_{22} 
\end{pmatrix},\\
I_X(u^*) =&
\begin{pmatrix}
L_{11} & L_{12} \\
L_{21} & L_{22} 
\end{pmatrix},
\end{align*}
where $K_{11}$ and $L_{11}$ are $d_1\times d_1$ matrices, and
$K_{22}$ and $L_{22}$ are $d_2\times d_2$ matrices.
We define the block elements of the inverse matrices as
\begin{align*}
I_{XY}(u^*)^{-1} =&
\begin{pmatrix}
\tilde{K}_{11} & \tilde{K}_{12}\\
\tilde{K}_{21} & \tilde{K}_{22}
\end{pmatrix},\\
I_X(u^*)^{-1} =&
\begin{pmatrix}
\tilde{L}_{11} & \tilde{L}_{12}\\
\tilde{L}_{21} & \tilde{L}_{22}
\end{pmatrix}.
\end{align*}
If $d_1=0$, we define the block matrix as
\begin{align*}
I_Z(u^*) =&
\begin{pmatrix}
A_{22} & A_{23} \\
A_{32} & A_{33}
\end{pmatrix},
\end{align*}
where $A_{22}$ is a $d_2\times d_2$ matrix
and $A_{33}$ is a $d_3\times d_3$ matrix.
Otherwise, we define it as
\begin{align*}
I_Z(u^*) =&
\begin{pmatrix}
0 & 0 & 0 \\
0 & A_{22} & A_{23} \\
0 & A_{32} & A_{33}
\end{pmatrix}.
\end{align*}
We also define the following inverse block matrix;
\begin{align*}
\begin{pmatrix}
A_{22} & A_{23} \\
A_{32} & A_{33}
\end{pmatrix}^{-1}
=&
\begin{pmatrix}
\tilde{A}_{22} & \tilde{A}_{23} \\
\tilde{A}_{32} & \tilde{A}_{33}
\end{pmatrix}.
\end{align*}

Theorem \ref{th:scondition} provides the unified asymptotic expression of the difference of the error functions $D(n)$ and $D_a(n)$
for all cases that gives
a sufficient condition for additional data to be effective.
\begin{theorem}
\label{th:scondition}
Let the block matrices of the Fisher information matrices $I_{XY}(u^*)$ and $I_X(u^*)$
be defined as above.
Let the eigenvalues of $\tilde{A}_{22}^{-1}\tilde{K}_{22}$ be $\lambda_1,\dots,\lambda_{d_2}$
and those of $\tilde{A}_{22}^{-1}\tilde{L}_{22}$ be $\mu_1,\dots,\mu_{d_2}$.
The asymptotic difference of the errors $D(n)$ and $D_a(n)$ is expressed as follows;
\begin{align*}
D(n) - D_a(n) =& \frac{1}{2}\ln \det (E_{d_2}+\alpha \tilde{A}_{22}^{-1}\tilde{L}_{22})
(E_{d_2}+\alpha \tilde{A}_{22}^{-1}\tilde{K}_{22})^{-1}\frac{1}{n} + o\bigg(\frac{1}{n}\bigg)\\
=& \frac{1}{2}\ln \bigg(\prod_{i=1}^{d_2}\frac{1+\alpha\mu_i}{1+\alpha\lambda_i}\bigg)\frac{1}{n} + o\bigg(\frac{1}{n}\bigg).
\end{align*}
The following condition is necessary and sufficient to ensure that the additional data set $D_a$ is effective:
\begin{align}
\prod_{i=1}^{d_2}(1+\alpha \mu_i)
>& \prod_{i=1}^{d_2}(1+\alpha \lambda_i). \label{eq:main_cond}
\end{align}
The following condition is sufficient; for all $i$,
\begin{align*}
\mu_i > \lambda_i.
\end{align*}
On the other hand,
if the coefficient $\ln \big(\prod_{i=1}^{d_2}\frac{1+\alpha \mu_i}{1+\alpha \lambda_i}\big)$ is negative,
the additional data degrade the accuracy.
\end{theorem}
It is obvious that the sufficient condition shows
\begin{align*}
1+\alpha \mu_i > 1 +\alpha \lambda_i > 0,
\end{align*}
which satisfies Eq.~(\ref{eq:main_cond}).

The proof of this theorem is presented in the next subsection,
and we will provide an interpretation of the theorem in Section \ref{sec:Dis}.
\subsection{Proof of Theorem \ref{th:scondition}}
We will first show the proof of the most general case, where $d_1,d_3>0$.
Since $d_1>0$ and $d_3>0$,
$I_{XY}(u^*)$, $I_X(u^*)$, and $I_Z(u^*)$ can be rewritten as
block matrices:
\begin{align*}
I_{XY}(u^*) =&
\begin{pmatrix}
K_{11} & K_{12} & 0 \\
K_{21} & K_{22} & 0 \\
0 & 0 & 0
\end{pmatrix},\\
I_X(u^*) =&
\begin{pmatrix}
L_{11} & L_{12} & 0 \\
L_{21} & L_{22} & 0 \\
0 & 0 & 0
\end{pmatrix},\\
I_Z(u^*) =&
\begin{pmatrix}
0 & 0 & 0 \\
0 & A_{22} & A_{23} \\
0 & A_{32} & A_{33}
\end{pmatrix}.
\end{align*}
According to Theorems \ref{th:asymD} and \ref{th:asymD_a},
\begin{align*}
D(n) - D_a(n) =& \frac{C_p}{2n} + o\bigg(\frac{1}{n}\bigg)\\
C_p =& \ln \det
\begin{pmatrix}
K_{11} & K_{12} \\
K_{21} & K_{22}
\end{pmatrix}
-\ln \det
\begin{pmatrix}
L_{11} & L_{12} \\
L_{21} & L_{22}
\end{pmatrix}\\
&-\ln \det \bigg\{
\begin{pmatrix}
K_{11} & K_{12} & 0 \\
K_{21} & K_{22} & 0 \\
0 & 0 & 0 
\end{pmatrix}
+\alpha
\begin{pmatrix}
0 & 0 & 0 \\
0 & A_{22} & A_{23} \\
0 & A_{32} & A_{33}
\end{pmatrix}\bigg\}\\
&+ \ln \det \bigg\{
\begin{pmatrix}
L_{11} & L_{12} & 0 \\
L_{21} & L_{22} & 0 \\
0 & 0 & 0
\end{pmatrix}
+\alpha
\begin{pmatrix}
0 & 0 & 0 \\
0 & A_{22} & A_{23} \\
0 & A_{32} & A_{33}
\end{pmatrix}\bigg\}.
\end{align*}
The coefficient can be rewritten as
\begin{align*}
C_p =&\ln \det
\begin{pmatrix}
K_{11} & K_{12} \\
K_{21} & K_{22}
\end{pmatrix}
-\ln \det
\begin{pmatrix}
L_{11} & L_{12} \\
L_{21} & L_{22}
\end{pmatrix}\\
&-\ln \det \bigg\{
\begin{pmatrix}
K_{11} & K_{12} & 0 \\
K_{21} & K_{22} & 0 \\
0 & 0 & 0 
\end{pmatrix}
\begin{pmatrix}
E_{d_1} & 0 & 0 \\
0 & \tilde{A}_{22} & \tilde{A}_{23} \\
0 & \tilde{A}_{32} & \tilde{A}_{33}
\end{pmatrix}\\
&+\alpha
\begin{pmatrix}
0 & 0 & 0 \\
0 & A_{22} & A_{23} \\
0 & A_{32} & A_{33}
\end{pmatrix}
\begin{pmatrix}
E_{d_1} & 0 & 0 \\
0 & \tilde{A}_{22} & \tilde{A}_{23} \\
0 & \tilde{A}_{32} & \tilde{A}_{33}
\end{pmatrix}\bigg\}\\
&+ \ln \det \bigg\{
\begin{pmatrix}
L_{11} & L_{12} & 0 \\
L_{21} & L_{22} & 0 \\
0 & 0 & 0
\end{pmatrix}
\begin{pmatrix}
E_{d_1} & 0 & 0 \\
0 & \tilde{A}_{22} & \tilde{A}_{23} \\
0 & \tilde{A}_{32} & \tilde{A}_{33}
\end{pmatrix}\\
&+\alpha
\begin{pmatrix}
0 & 0 & 0 \\
0 & A_{22} & A_{23} \\
0 & A_{32} & A_{33}
\end{pmatrix}
\begin{pmatrix}
E_{d_1} & 0 & 0 \\
0 & \tilde{A}_{22} & \tilde{A}_{23} \\
0 & \tilde{A}_{32} & \tilde{A}_{33}
\end{pmatrix}\bigg\}\\
=& \ln \det
\begin{pmatrix}
K_{11} & K_{12} \\
K_{21} & K_{22}
\end{pmatrix}
-\ln \det
\begin{pmatrix}
L_{11} & L_{12} \\
L_{21} & L_{22}
\end{pmatrix}\\
&-\ln \det
\begin{pmatrix}
K_{11} & K_{12}\tilde{A}_{22} & K_{12}\tilde{A}_{23} \\
K_{21} & K_{22}\tilde{A}_{22}+\alpha E_{d_2} & K_{22}\tilde{A}_{23} \\
0 & 0 & \alpha E_{d_3}
\end{pmatrix}\\
&+\ln \det
\begin{pmatrix}
L_{11} & L_{12}\tilde{A}_{22} & L_{12}\tilde{A}_{23} \\
L_{21} & L_{22}\tilde{A}_{22} + \alpha E_{d_2} & L_{22}\tilde{A}_{23} \\
0 & 0 & \alpha E_{d_3}
\end{pmatrix}\\
=& \ln \det
\begin{pmatrix}
K_{11} & K_{12} \\
K_{21} & K_{22}
\end{pmatrix}
-\ln \det
\begin{pmatrix}
L_{11} & L_{12} \\
L_{21} & L_{22}
\end{pmatrix}\\
&-\ln \det
\begin{pmatrix}
K_{11} & K_{12}\tilde{A}_{22} \\
K_{21} & K_{22}\tilde{A}_{22}+\alpha E_{d_2}
\end{pmatrix}
+ \ln \det
\begin{pmatrix}
L_{11} & L_{12}\tilde{A}_{22} \\
L_{21} & L_{22}\tilde{A}_{22}+\alpha E_{d_2}
\end{pmatrix}\\
=& \ln \det
\begin{pmatrix}
K_{11} & K_{12} \\
K_{21} & K_{22}
\end{pmatrix}
-\ln \det
\begin{pmatrix}
L_{11} & L_{12} \\
L_{21} & L_{22}
\end{pmatrix}\\
&-\ln \det \bigg\{
\begin{pmatrix}
K_{11} & K_{12} \\
K_{21} & K_{22}
\end{pmatrix}
\begin{pmatrix}
E_{d_1} & 0 \\
0 & \tilde{A}_{22}
\end{pmatrix}
+\alpha
\begin{pmatrix}
0 & 0 \\
0 & E_{d_2}
\end{pmatrix}\bigg\}\\
&+\ln \det \bigg\{
\begin{pmatrix}
L_{11} & L_{12} \\
L_{21} & L_{22}
\end{pmatrix}
\begin{pmatrix}
E_{d_1} & 0 \\
0 & \tilde{A}_{22}
\end{pmatrix}
+\alpha
\begin{pmatrix}
0 & 0 \\
0 & E_{d_2}
\end{pmatrix}\bigg\}.
\end{align*}
Finally, we obtain the following expression for the coefficient:
\begin{align*}
C_p =& -\ln \det \bigg\{
\begin{pmatrix}
K_{11} & K_{12} \\
K_{21} & K_{22}
\end{pmatrix}
+\alpha
\begin{pmatrix}
0 & 0 \\
0 & \tilde{A}_{22}^{-1}
\end{pmatrix}\bigg\}\\
&+ \ln \det \bigg\{
\begin{pmatrix}
L_{11} & L_{12} \\
L_{21} & L_{22}
\end{pmatrix}
+\alpha
\begin{pmatrix}
0 & 0 \\
0 & \tilde{A}_{22}^{-1}
\end{pmatrix}\bigg\}\\
&+\ln \det
\begin{pmatrix}
K_{11} & K_{12} \\
K_{21} & K_{22}
\end{pmatrix}
\begin{pmatrix}
L_{11} & L_{12} \\
L_{21} & L_{22}
\end{pmatrix}^{-1}\\
= & -\ln \det \bigg\{
\begin{pmatrix}
E_{d_1} & 0 \\
0 & E_{d_2} 
\end{pmatrix}
+ \alpha
\begin{pmatrix}
0 & 0 \\
\tilde{A}_{22}^{-1}\tilde{K}_{21} & \tilde{A}_{22}^{-1}\tilde{K}_{22}
\end{pmatrix}\bigg\}\\
&+ \ln \det \bigg\{
\begin{pmatrix}
E_{d_1} & 0 \\
0 & E_{d_2}
\end{pmatrix}
+ \alpha
\begin{pmatrix}
0 & 0 \\
\tilde{A}_{22}^{-1}\tilde{L}_{21} & \tilde{A}_{22}^{-1}\tilde{L}_{22}
\end{pmatrix}\bigg\}\\
=& -\ln \det
\begin{pmatrix}
E_{d_1} & 0 \\
0 & \alpha \tilde{A}_{22}^{-1}\tilde{K}_{22} + E_{d_2}
\end{pmatrix}
+ \ln \det
\begin{pmatrix}
E_{d_1} & 0 \\
0 & \alpha \tilde{A}_{22}^{-1}\tilde{L}_{22} + E_{d_2}
\end{pmatrix}\\
=& -\ln \det (\alpha \tilde{A}_{22}^{-1}\tilde{K}_{22} + E_{d_2}) 
+ \ln \det (\alpha \tilde{A}_{22}^{-1}\tilde{L}_{22} + E_{d_2}).
\end{align*}
Using the eigenvalues, we obtain
\begin{align*}
C_p =& -\ln \prod_{i=1}^{d_2}(1+\alpha \lambda_i) + \ln \prod_{i=1}^{d_2}(1+\alpha \mu_i),
\end{align*}
which proves the last form of the difference.

By omitting the unnecessary parts of the block matrices, it is easy to prove the other cases.
\subsection{Application to Selection of Additional Data Sets}
Since we have obtained the asymptotic form of the difference between the errors with and without the additional data,
we apply the result to the data analysis.
Assume that there are some candidates of additional data sets,
and that there are the models to express their data source after the thorough model selection.
We compare the effects of the data sets on the accuracy of the clustering and select the optimal set to improve the error.
In the same way as the model selection,
the asymptotic difference of the error can be used for the criterion \citep{Akaike}.

In the Bayes clustering, it is known that the conjugate prior reduces the computational complexity and
eliminates redundant dimensions of the parameter in the mixture model \citep{Yamazaki2016,Yamazaki14c}.
Accordingly, we assume that the density function $f$ of the components in the model can attain the one in the true distribution
and there is no redundant components, using the conjugate prior.

Let $D_a^i$ be the $i$th additional data set, where the number of data is $\alpha_i n$,
and $\hat{u}^i$ be the maximum a posteriori (MAP) estimator defined by
\begin{align}
\hat{u}^i =& \arg\max_u \prod_{j=1}^np_i(x_j,y_j|u)\prod_{j=n+1}^{n+\alpha n}p_a(z_j|u)\varphi(u). \label{eq:MAP}
\end{align}
Note that the dimension of the parameter $u$, its prior $\varphi(u)$
and the model for the additional part $p_a(z|u)$ also depend on the data set $D_a^i$
though we omit the suffix $i$ of them to simplify the notations.
The empirical Fisher information matrices are given by
\begin{align*}
\{I_{XY}(\hat{u}^i)\}_{jk} =& \frac{1}{n'}\sum_{l=1}^{n'} \bigg[ \frac{\partial \ln p_i(x_l,y_l|\hat{u}^i)}{\partial u_j}
\frac{\partial \ln p_i(x_l,y_l|\hat{u}^i)}{\partial u_k} \bigg],\\
\{I_X(\hat{u}^i)\}_{jk} =& \frac{1}{n'}\sum_{l=1}^{n'} \bigg[ \frac{\partial \ln p_i(x_l|\hat{u}^i)}{\partial u_j}
\frac{\partial \ln p_i(x_l|\hat{u}^i)}{\partial u_k} \bigg],\\
\{I_Z(\hat{u}^i)\}_{jk} =& \frac{1}{\alpha_i n'}\sum_{l=n'+1}^{n'+\alpha_i n'} \bigg[ \frac{\partial \ln p_a(z_l|\hat{u}^i)}{\partial u_j}
\frac{\partial \ln p_a(z_l|\hat{u}^i)}{\partial u_k} \bigg].
\end{align*}
The block inverse matrices $\tilde{K}_{22}(\hat{u}^i)$, $\tilde{L}_{22}(\hat{u}^i)$ and $\tilde{A}_{22}(\hat{u}^i)$
are defined on the basis of the estimator.
Let the eigenvalues of $\tilde{A}_{22}(\hat{u}^i)^{-1}\tilde{K}_{22}(\hat{u}^i)$ be $\lambda_1(\hat{u}^i),\dots,\lambda_{d_2}(\hat{u}^i)$
and those of $\tilde{A}_{22}(\hat{u}^i)^{-1}\tilde{L}_{22}(\hat{u}^i)$ be $\mu_1(\hat{u}^i),\dots,\mu_{d_2}(\hat{u}^i)$.
Then, we select the additional data set maximizing the factor
\begin{align*}
\mathrm{IC}(\hat{u}^i) =& \prod_{j=1}^{d_2} \frac{1+\alpha_i \mu_j(\hat{u}^i)}{1+\alpha_i \lambda_j(\hat{u}^i)}.
\end{align*}
Since the above matrices converge to the Fisher information matrices $I_{XY}(u^*)$, $I_X(u^*)$ and $I_Z(u^*)$, respectively,
this factor converges to the coefficient of the asymptotic difference of the errors in Theorem \ref{th:scondition}.
Therefore, maximizing this factor allows us to select the optimal data set improving the asymptotic accuracy.
\subsection{Numerical Computation of IC}
In this subsection, we compute the IC value with a simple model setting
to evaluate how the asymptotic criterion behaves in non-asymptotic situations.
Let us consider a Gaussian mixture model defined by
\begin{align*}
p(x|w) =& a_1 \mathcal{N}(x|b_1) + a_2 \mathcal{N}(x|b_2),
\end{align*}
where $x\in R^1$ and $\mathcal{N}(x|b)$ is one-dimensional Gaussian distribution
with the mean parameter $b \in R^1$ and the fixed variance $\sigma^2=1$. 
According to the definition, $a_2=1-a_1$.
The parameter is expressed as $w=(a_1,b_1,b_2)^\top$.

\begin{figure}[th]
\centering
\includegraphics[width=0.8\columnwidth]{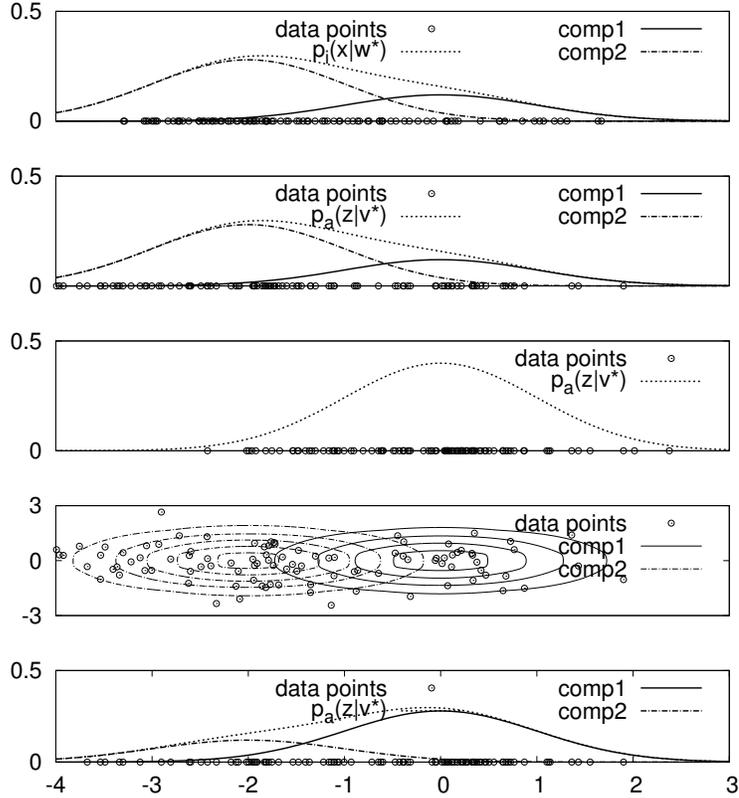}
\caption{From the top panel, the data points and the shapes of their distributions for $D_i$, $D^1_a$, $D^2_a$, $D^3_a$ and $D^4_a$, respectively.
The notations 'comp1' and 'comp2' are the distributions of the first and the second components of the mixture, respectively.}
\label{fig:dist}
\end{figure}

In the numerical experiments, the true parameter of $p_i(x|w^*)$ was $w^*=(0.3,0,-2)^\top$,
which implies $a^*_1=0.3$, $b^*_1=0$ and $b^*_2=-2$.
There were five cases on the number of the initial data: $n=10,50,100,500,1000$.
The following four additional dat sets were compared;
the first additional data set $D^1_a$ was generated by
\begin{align*}
p_a(z|v^*) =& a_1^* \mathcal{N}(x|b^*_1)+(1-a^*_1)\mathcal{N}(x|b^*_2),
\end{align*}
where $p_i(x|w^*)=p_a(z|v^*)$
corresponding to Example \ref{ex:TypeIId} ($d_1=d_3=0$).
The second additional data set $D^2_a$ was generated by
\begin{align*}
p_a(z|v^*) =& \mathcal{N}(x|b^*_1),
\end{align*}
where all data were from the first component and their labels were given,
i.e., $z=(x,y=1)^\top$
corresponding to Example \ref{ex:one_class} ($d_1>0,d_3=0$).
The third additional data set $D^3_a$ was generated by
\begin{align*}
p_a(z|v^*) =& a^*_1\mathcal{N}(x|b^*_1)\mathcal{N}(x'|c^*_1)+(1-a^*_1)\mathcal{N}(x|b^*_2)\mathcal{N}(x'|c^*_2),
\end{align*}
where the data had the additional dimension $x'\in R^1$, i.e. $z=(x,x')^\top$  and $c^*_1=c^*_2=0$
corresponding to Example \ref{ex:add_feature} ($d_1=0,d_3>0$).
The last additional data set $D^4_a$ was generated by
\begin{align*}
p_a(z|v^*) =& c^*_1 \mathcal{N}(x|b^*_1) +(1-c^*_1)\mathcal{N}(x|b^*_2),
\end{align*}
where the class prior $a^*_1=0.3$ had been changed into $c^*_1=0.7$
corresponding to Example \ref{ex:CPchange} ($d_1>0,d_3>0$).
The number of the additional data was the same as that of the initial data for all additional data sets,
which means $\alpha=1.0$.
Fig.\ref{fig:dist} shows the data points and the shapes of $p_i(x|w^*)$ and $p_a(z|v^*)$.

\begin{figure}[th]
\centering
\includegraphics[width=\columnwidth]{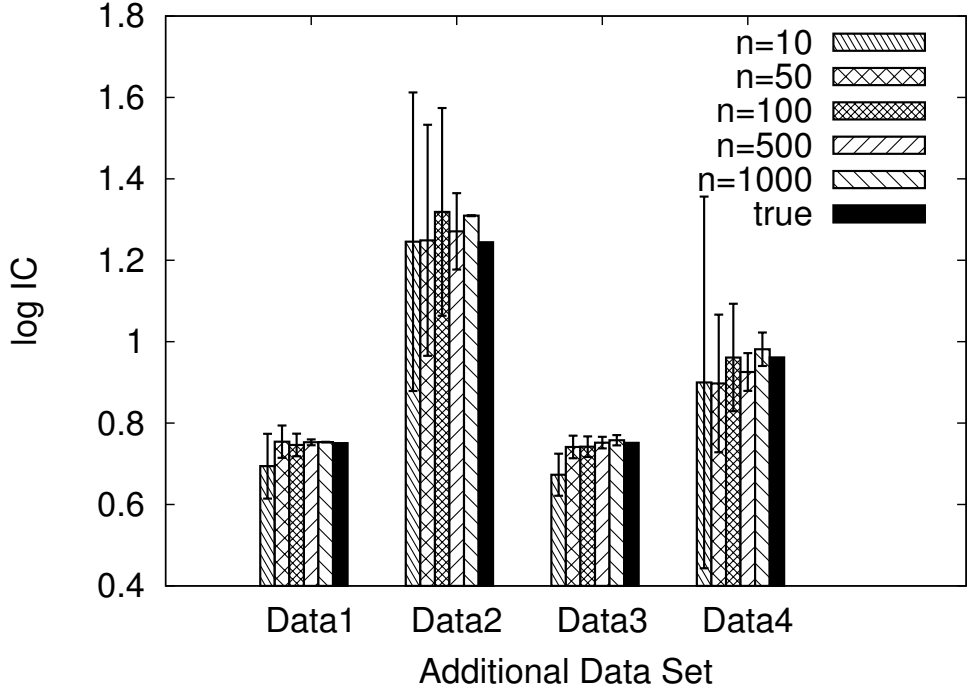}
\caption{The IC values: the horizontal axis shows the additional data sets 
and the vertical one shows the IC value in the logarithmic scale.}
\label{fig:IC}
\end{figure}

Fig.\ref{fig:IC} shows the IC value of each additional data set.
The horizontal axis shows the additional data sets;
'Data $i$' stands for the $i$th additional data set $D^i_a$.
The vertical one shows the log-scale IC values.
The 'true' values are the results with the true parameter,
and the others have the average value and the standard deviation.
When the parameters were estimated,
the initial data set and the additional data sets were generated for five times in each data size $n$.
The MAP estimators $\hat{u}^i$ for $i=1,2,\dots,4$ were calculated based on Eq.~(\ref{eq:MAP}).
In order to compute the Fisher information matrices such as $I_{XY}(\hat{u}^i)$, $I_X(\hat{u}^i)$ and $I_Z(\hat{u}^i)$,
500,000 data points were regenerated from $p_i(x,y|\hat{u}^i)$ and $p_a(z|\hat{u}^i)$, i.e., $n'=500,000$.
Note that $n$ is the number of data to obtain the MAP estimator while $n'$ is that to calculate the IC values.
This means that the estimator was obtained in non asymptotic situations when $n$ was small.

As shown in Fig.\ref{fig:IC}, the approximation of the IC value becomes reliable with the growth of the number of data $n$.
Considering the selection of the additional data,
the second data set $D^2_a$ was the most informative on the labels in the experimental setting.
Since the differences from the other data sets were large,
the selected data set according to the IC values was the second one in each trial;
\begin{table}[th]
\centering
\caption{The log IC values in each trial}
\label{tab:trials}
\begin{tabular}{|l|c|c|c|c|}
\hline
$n=10$&Data1&Data2&Data3&Data4\\
\hline \hline
true & 0.750728 & \bf{1.244250} & 0.751362 & 0.961213 \\
\hline
Trial 1 & 0.762980 & \bf{0.796713} & 0.747058 & 0.599663 \\
\hline
Trial 2 & 0.571954 & \bf{0.891112} & 0.633154 & 0.478174 \\
\hline
Trial 3 & 0.641944 & \bf{1.317793} & 0.637458 & 0.556322 \\
\hline
Trial 4 & 0.703079 & \bf{1.802498} & 0.623486 & 1.643882 \\
\hline
Trial 5 & 0.790424 & \bf{1.419697} & 0.724202 & 1.220538 \\
\hline
\end{tabular}
\end{table}
Table \ref{tab:trials} shows the IC values of the five trials in $n=10$
and the largest ones are displayed in bold.
Even in the smallest size of data, the second data set indicated the largest IC value
while the magnitude relation among the remaining sets depended on the given data.

The first and third data sets had stable approximation for the IC values
though they did not have large values.
On the other hand the second and fourth data sets were rather large deviations but informative.
The difference between these two groups was the setting of the mixing ratio;
$D^1_a$ and $D^3_a$ had the same ratio of the labels as the initial data set,
whereas $D^2_a$ and $D^4_a$ had the different one.
This implies that changing the mixing ratio in the additional data improves the accuracy of the estimation of the labels
and at the same time it makes the additional-data selection unstable.

In the experimental settings, the first component $\mathcal{N}(x|b^*_1=0)$,
which was the right component in Fig.~\ref{fig:dist} had the smaller number of data
compared with the second one due to the mixing ratio $a^*_1=0.3$.
The second data set $D^2_a$ intensively provided the data from the first component.
The fourth data set $D^4_a$ also complemented the first component though the amount of the addional data was rather small
and this different appeared in the magnitude relation of the IC values.
The complement reducing the uncertainty of the initial data set increased the IC values.

As for the numerical evaluation of the performance,
it is important to emphasize that rigorous calculation of the accuracy defined by the KL divergence Eq.~\ref{eq:def_Da} is not straightforward
in non-asymptotic situations.
As shown in the proof of Theorem \ref{th:asymD_a}, the error function is divided into two free energy functions:
\begin{align*}
D_a(n) =& \frac{1}{n}\bigg\{ F_{XY}(n) - F_X(n) \bigg\},
\end{align*}
which holds in any data size $n$.
If we do not rely on the asymptotic form of $D_a(n)$,
we need to calculate the values of these free energy functions based on the experimental results.
However, it has been known that there is no exact computational method for the latter function $F_X(n)$
and some approximation is necessary such as the Laplace approximation, the variational Bayes method and the MCMC method.
The energy function includes the integral of the parameter, which requires analytic calculation or parameter sampling from the posterior distribution.
The Laplace approximation provides the same results as the asymptotic form.
The variational Bayes method has different energy value since the variational free energy is the lower bound of the original energy function \citep{Attias99}.
The sophisticated MCMC method such as the exchange Monte Carlo method \citep{Swendsen1986,Ogata} is required for the hierarchical probabilistic models such as the mixture model \citep{Nagata08}.
Therefore, the calculation of the accuracy from the experimental results always includes the discrepancy between $F_X(n)$ and its approximation,
which means that we evaluate not only the error function itself but also the approximation method of the integral of the parameter.
The asymptotic form is so far the only result to calculate the rigorous error value in the sense of no approximation.
This is a different property from other error function;
for example, the 0-1 loss is enabled to be computed when the labels are given in the numerical evaluation.
To develop an accurate and efficient calculation method of the error function $D_a(n)$ is one of our challenging future studies.
\section{Discussion}
\label{sec:Dis}
In this section, we begin by summarizing the meaning of the Fisher information matrix, and then we provide an interpretation
of Theorem \ref{th:scondition}.
Next, we compare theorem with the result in the former study.
Last, we show that Bayesian transfer learning,
which uses the posterior distribution of the parameter obtained by the reference/additional data
as the prior distribution for the target domain,
has the same formulation as our estimation, which is expressed as $p(Y|X,D_a)$.
\subsection{Interpretation of Theorem \ref{th:scondition}}
Before considering the interpretation,
let us first introduce one of the meanings of the Fisher information matrix.
We define the following maximum-likelihood estimators:
\begin{align*}
\hat{u}_{XY} =& \arg \max_u \prod_{j=1}^n p_i(x_j,y_j|u),\\
\hat{u}_X =& \arg \max_u \prod_{j=1}^n p_i(x_j|u).
\end{align*}
Here, we assume
$\hat{u}_{XY}\rightarrow u^*$ and $\hat{u}_X \rightarrow u^*$ for sufficiently large $n$,
which is satisfied in Theorem \ref{th:scondition}.
This assumption means that $\hat{u}_X$ is located in the neighborhood of $\hat{u}_{XY}$.
Since these estimators are unbiased,
their distributions can be expressed as 
$\mathcal{N}(\hat{u}_{XY}|u^*,\frac{1}{n}I_{XY}(u^*)^{-1})$ and
$\mathcal{N}(\hat{u}_X|u^*,\frac{1}{n}I_X(u^*)^{-1})$ \citep{vdVaart98},
where $\mathcal{N}(\cdot|\mu,\Sigma)$ is a normal distribution with mean $\mu$ and variance-covariance matrix $\Sigma$.
The Fisher information matrices $I_{XY}(u^*)$ and $I_X(u^*)$ are the precision matrices
of the distribution of the estimators $\hat{u}_{XY}$ and $\hat{u}_X$, respectively.
In this sense, the Fisher information matrix shows the precision of the model.

Now, we focus on the leading term of the asymptotic difference.
The factor in the term is rewritten as
\begin{align}
&\ln \det (E_{d_2}+\alpha \tilde{A}_{22}^{-1}\tilde{L}_{22})
(E_{d_2}+\alpha \tilde{A}_{22}^{-1}\tilde{K}_{22})^{-1}\nonumber\\
&= \ln \det \tilde{K}_{22}^{-1}(\tilde{L}_{22}^{-1})^{-1} - \ln \det (\tilde{K}_{22}^{-1}+\alpha \tilde{A}_{22}^{-1})
(\tilde{L}_{22}^{-1}+\alpha \tilde{A}_{22}^{-1})^{-1}.\label{eq:intp}
\end{align}
The matrices $\tilde{A}_{22}^{-1}$, $\tilde{K}_{22}^{-1}$ and $\tilde{L}_{22}^{-1}$
correspond to the precision of the models $p_a(z|u)$, $p_i(x,y|u)$ and $p_i(x|u)$, respectively.
We can find that the factor consists of the block matrices with the same suffix;
the asymptotic difference is determined by the precision of the common part of parameters $w$ and $v$.
This implies that the result of the parameter estimation for the individual dimension does not have
direct connection to improvement of the clustering accuracy.

When the additional data are informative and $\tilde{A}_{22}^{-1}$ is dominant in the second term of the right-hand side in Eq.~(\ref{eq:intp}),
the product of the matrices is close to the unit matrix,
which means that the second term vanishes and the asymptotic difference is maximized.
On the other hand, the difference is close to zero
if the effect of $\tilde{A}_{22}^{-1}$ is small and the second term is almost equivalent to the first one.
Moreover, the amount of additional data also has the same function;
the large $\alpha$ makes the difference large.

Next, let us now consider the difference of the eigenvalues $\lambda_i$ and $\mu_i$,
which appears in the sufficient condition of the theorem.
According to the definitions,
\begin{align*}
\ln \det \tilde{A}_{22}^{-1} - \ln \det \tilde{K}_{22}^{-1} =& \ln \prod_{j=1}^{d_2} \lambda_j,\\
\ln \det \tilde{A}_{22}^{-1} - \ln \det \tilde{L}_{22}^{-1} =& \ln \prod_{j=1}^{d_2} \mu_j.
\end{align*}
Comparison of these eigenvalues is
to find the difference between $\tilde{K}_{22}^{-1}$ and $\tilde{L}_{22}^{-1}$,
which are the precisions with and without the label, respectively.
Therefore, the sufficient condition requires that the precision of the initial model is improved by obtaining the label $y_i$
in each direction of the eigenspace.
\subsection{Comparison with the Result of the Former Study}
We compare Theorem \ref{th:scondition} with the result of the former study in \citet{Yamazaki15c},
which has focused on Example \ref{ex:CPchange} and revealed the clustering accuracy when the class prior changes.
From the technical point of view, the Fisher information matrices $I_{XY}$ and $I_Z$ of this former study
are the diagonal block matrix, where $K_{12}=0$, $K_{21}=0$, $A_{23}=0$ and $A_{32}=0$,
since the additional data are restricted to labeled data
and then the parameters of the class priors are independent of those of the components.
This independence is expressed as the product form of the model
such as $b_y$ and $c_y$ in $p_a(x,y|u)=c_yf(x|b_y)$ \citep{Yamazaki15c},
and implies the constraint $K_{22}=A_{22}$,
which makes the analysis straightforward.
For example, the inverse matrices are written as
\begin{align*}
I_{XY}^{-1}=&
\begin{pmatrix}
K_{11} & 0 \\
0 & K_{22}
\end{pmatrix}^{-1}
=
\begin{pmatrix}
K_{11}^{-1} & 0\\
0 & K_{22}^{-1}
\end{pmatrix}\\
I_{Z}^{-1}=&
\begin{pmatrix}
A_{22} & 0 \\
0 & A_{33}
\end{pmatrix}^{-1}
=
\begin{pmatrix}
A_{22}^{-1} & 0\\
0 & A_{33}^{-1}
\end{pmatrix},
\end{align*}
and it is easy to distinguish the nuisance parts for the accuracy such as $K_{11}^{-1}$ and $A_{33}^{-1}$
when the determinant of their multiplied matrices are calculated.
Moreover, the condition for the effective additional data is simply described by $\mu_i>1$
since the inverse block matrices are equivalent $\tilde{K}_{22}=K_{22}^{-1}=A_{22}^{-1}=\tilde{A}_{22}$
and the eigenvalues of the matrix $\tilde{A}_{22}^{-1}\tilde{K}_{22}=E_{d_2}$ are all one.

The present paper generalizes the case and removes the constraint, i.e., $I_{XY}$ and $I_Z$ can be non-diagonal block matrices.
This generalization allows the theorem to be applied to many practical situations.
One of the typical examples is that the additional data can be unlabeled,
where the additional part of the model is expressed as $p_a(x|u)=\sum_y c_yf(x|b_y)$.
Another example is that there is no constraint of the product form of the model.
Let us consider the following case;
a part of dimensions in the parameter of the component also changes
between the initial and the additional models. For example,
\begin{align*}
p_i(x|u) =& a_yf(x|b_y),\\
p_a(x,y|u) =& c_y f(x|\bar{b}_y),
\end{align*}
where $\bar{b}_y$ is defined by
\begin{align*}
\bar{b}_{yi} =& b_{yi} \;\; (1\le i \le M),\\
\bar{b}_{yi} \neq& b_{yi} \;\; (M+1 \le i \le d_c)
\end{align*}
for $M<d_c$.
Even for this complicated change between the initial and the additional data,
Theorem \ref{th:scondition} is available.
\subsection{Bayesian Transfer Learning}
In the Bayes method,
one way to transfer the prior knowledge to the target task is to optimize the prior distribution of the model parameter such that it maximizes the expression of knowledge.
Let us regard the additional data set $D_a$ as the source of the prior knowledge.
The posterior distribution of the parameter on the prior knowledge is given by
\begin{align*}
p(v|D_a) =& \frac{1}{Z_a}\prod_{i=n+1}^{n+\alpha n}p_a(z_i|v)\varphi(v),
\end{align*}
where $Z_a$ is the normalizing constant, and $\varphi(v)$ is a prior distribution.
On the other hand, the estimate of the target data can be expressed as
\begin{align*}
p(Y|X) =& \frac{\int \prod_{j=1}^n p_i(x_j,y_j|w)\varphi(w)dw}{\int \prod_{j=1}^n p_i(x_j|w)\varphi(w)dw},
\end{align*}
where $\varphi(w)$ is the prior distribution of the target task.
To transfer the knowledge based on $D_a$ to this task,
we replace $\varphi(w)$ with the posterior $p(v|D_a)$.
Since there is a dimensional gap between the parameter spaces of $w$ and $v$,
we consider the extended space denoted by $u$, which was defined by the functions $\psi_i(w)$ and $\psi_a(v)$ in Section \ref{sec:Def_Data}:
\begin{align*}
p_t(Y|X) =& \frac{\int \prod_{j=1}^n p_i(x_j,y_j|u)p(u|D_a)du}{\int \prod_{j=1}^n p_i(x_j|u)p(u|D_a)du},
\end{align*}
where $p(u|D_a)$ is defined as
\begin{align*}
p(u|D_a) =& \varphi(u_r)p(\psi_a(v)|D_a),\\
p(\psi_a(v)|D_a) =& \frac{1}{Z_a}\prod_{i=n+1}^{n+\alpha n}p_a(z_i|\psi_a(v))\varphi(\psi_a(v)).
\end{align*}
We use the notation $u_r=(u_1,\dots,u_{d_1})$, where $u=(u_r,\psi_a(v))$.
Then, it can be easily found that
\begin{align*}
p_t(Y|X) =& \frac{\int \prod_{j=1}^n p_i(x_j,y_j|u)\varphi(u_r)p(\psi_a(v)|D_a)du}
{\int \prod_{j=1}^n p_i(x_j|u)\varphi(u_r)p(\psi_a(v)|D_a)du}\\
=& \frac{\int \prod_{j=1}^n p_i(x_j,y_j|u)\prod_{i=n+1}^{n+\alpha n}p_a(z_i|u)\varphi_t(u)du}
{\int \prod_{j=1}^n p_i(x_j|u)\prod_{i=n+1}^{n+\alpha n}p_a(z_i|u)\varphi_t(u)du},
\end{align*}
where $\varphi_t(u)=\varphi(u_r)\varphi(\psi_a(v))$.
The last line corresponds to $p(Y|X,D_a)$, which was defined in Section \ref{sec:Def_Data}.

This equivalence relation claims that the estimation of the target data does not require all of the additional data;
sample parameters taken from the posterior $p(\psi_a(v)|D_a)$ are sufficient for calculating $p_t(Y|X)$.
Let $V=\{ \psi_a(v_1),\dots,\psi_a(v_{n_p})\}$ be the sample of the parameter from the posterior distribution.
The distribution of the estimate can be written as
\begin{align*}
p_t(Y|X) = p(Y|X,D_a) \propto& \int \prod_{j=1}^n p_i(x_j,y_j|u)\varphi(u_r)p(\psi_a(v)|D_a)du\\
=& \int \prod_{j=1}^n p_i(x_j,y_j|u)\varphi(u_r)du_r p(\psi_a(v))d\psi_a(v)\\
\simeq & \frac{1}{n_p}\sum_{j=1}^{n_p} \int \prod_{k=1}^n p_i(x_k,y_k|u_r,\psi_a(v_j))\varphi(u_r)du_r.
\end{align*}
Based on the last line, we can use $V$ instead of $D_a$ to estimate $Y$.
%
%
%
\section{Conclusion}
\label{sec:Con}
We analyzed the effect of using additional data on the accuracy of estimating the latent variable.
There is a trade-off between the amount of data used and the complexity of the model.
According to the asymptotic analysis, 
the difference between the errors with and without the additional data has been quantitatively revealed,
and the condition, in which the advantages of using additional data outweigh the disadvantages of increased complexity,
has been derived.
\section*{Acknowledgments}
This research was partially supported by a research grant by the Support Center for Advanced Telecommunications Technology Research Foundation
and by KAKENHI 15K00299.
%
%
\vskip 0.2in
\bibliography{LearningTheory}
\bibliographystyle{natbib}
\end{document}